\documentclass[pdflatex,sn-mathphys-num,iicol]{sn-jnl}  

\usepackage{graphicx}%
\usepackage{multirow}%
\usepackage{amsmath,amssymb,amsfonts}%
\usepackage{amsthm}%
\usepackage{mathrsfs}%
\usepackage[title]{appendix}%
\usepackage{xcolor}%
\usepackage{textcomp}%
\usepackage{manyfoot}%
\usepackage{booktabs}%
\usepackage{algorithm}%
\usepackage{algorithmicx}%
\usepackage{algpseudocode}%
\usepackage{listings}%

\usepackage{adjustbox}
\usepackage{subcaption}
\usepackage{arydshln}
\usepackage{multicol}
\usepackage{tabularx}
\usepackage[normalem]{ulem}

\definecolor{ngreen}{RGB}{17, 173, 30}

\theoremstyle{thmstyleone}%
\theoremstyle{thmstyletwo}%

\theoremstyle{thmstylethree}%

\begin{document}

\title{Enhancing Video Large Language Models with Structured Multi-Video Collaborative Reasoning}








\author[1,2]{\fnm{Zhihao} \sur{He}}\email{ziho\_he@sjtu.edu.cn}

\author[1]{\fnm{Tianyao} \sur{He}}

\author[1]{\fnm{Yun} \sur{Xu}}

\author[1]{\fnm{Tieyuan} \sur{Chen}}

\author[1]{\fnm{Huabin} \sur{Liu}}

\author[1]{\fnm{Chaofan} \sur{Gan}}

\author[2,3]{\fnm{Zuxuan} \sur{Wu}}

\author*[1]{\fnm{Weiyao} \sur{Lin}}\email{wylin@sjtu.edu.cn}

\affil[1]{\orgname{Shanghai Jiao Tong University}}

\affil[2]{\orgname{Shanghai Innovation Institute}}

\affil[3]{\orgname{Fudan University}}


\abstract{
    Despite the prosperity of the video language model, the current pursuit of comprehensive video reasoning is thwarted by the inherent spatio-temporal incompleteness within individual videos, resulting in hallucinations and inaccuracies.
    A promising solution is to augment the reasoning performance with multiple related videos. However, video tokens are numerous and contain redundant information, so directly feeding the relevant video data into a large language model to enhance responses could be counterproductive.
    To address this challenge, we propose a multi-video collaborative framework for video language models. For efficient and flexible video representation, we establish a Video Structuring Module to represent the video's knowledge as a spatio-temporal graph. Based on the structured video representation, we design the Graph Fusion Module to fuse the structured knowledge and valuable information from related videos into the augmented graph node tokens. Finally, we construct an elaborate multi-video structured prompt to integrate the graph, visual, and textual tokens as the input to the large language model. Extensive experiments substantiate the effectiveness of our framework, showcasing its potential as a promising avenue for advancing video language models.
    Code is open-sourced at \url{https://github.com/ziHoHe/SMV-CR}.
}




\maketitle

\begin{figure}
    \centering
    \includegraphics[width=1.0\linewidth]{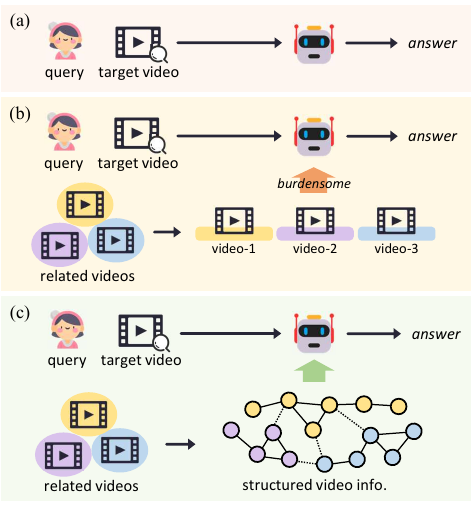}
    \caption{Video question-answering pipeline under different video collaboration strategies. (a) \textbf{Single-video reasoning pipeline}; (b) \textbf{Direct multi-video collaboration pipeline:} concatenate multiple video's visual tokens, which is burdensome; (c) \textbf{Structured multi-video collaboration pipeline} (ours).}
    \label{fig:head_1}
\end{figure}

\begin{figure*}
    \centering
    \includegraphics[width=1.0\linewidth]{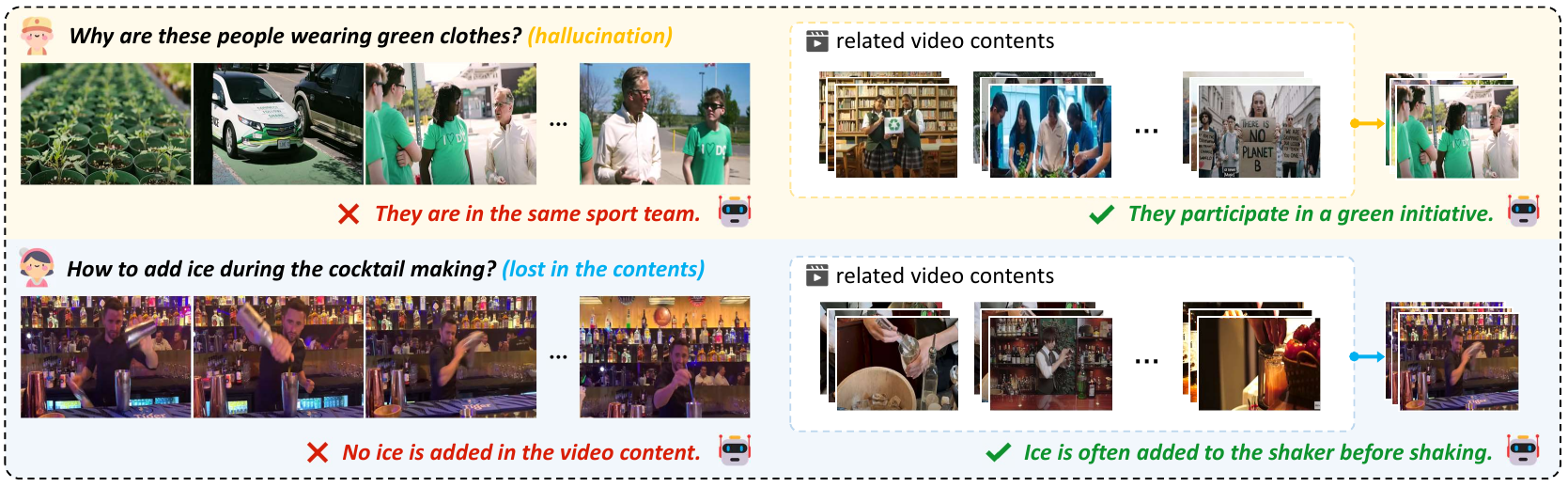}
    \caption{\textbf{Video question answering examples from video language models.} \textit{Single-video reasoning (Left):} In video-1, the environmental visual cues are hard for the model to perceive, leading to a 'sports team' hallucination based on only the textual query and linguistic priors. In Video 2, ice-related visuals are missing, and limited bartending knowledge causes the model to skip the question. \textit{Multi-video reasoning (Right):} Introducing relevant videos allows the model to complete and summarize domain-specific knowledge (such as environmental protection or bartending in this case), leading to more reliable and accurate answers.}
    \label{fig:single_vs_multiple}
\end{figure*}

\section{Introduction}
\label{sec:intro}
    The significant advancement in Large Language Models (LLMs) gives rise to the prosperity of video language models (VLMs). Equipped with LLMs' strong general knowledge, video language models showcase an impressive potential to comprehend complex video contents~\cite{li2023llama,yang2023vid2seq,zhang2023video,li2023videochat,maaz2023video,lin2024video,luo2023valley}.
    These models effectively bridge the semantic chasm between video and language, facilitating a nuanced integration of visual perception with language processing.
    
    However, it remains an elusive goal to achieve reliable and comprehensive video reasoning. Specifically, the spatio-temporal knowledge captured from an individual video is often incomplete due to the limited \textit{visual perception} and \textit{comprehending ability}. Factors hindering visual perception include sparse sampling~\cite{liu2022task,yang2022tubedetr,yu2021learning,piergiovanni2023rethinking}, spatial occlusion~\cite{jin2022otpose,dong2023erasing}, and perspective switching~\cite{shot_switching1,shot_switching2,perspective_shifts1,perspective_shifts2}). The comprehending ability is mainly constrained by the complexity of video content~\cite{xia2023structured,ji2020action} and limited effective context length~\cite{liu2024lost,hsieh2024ruler}. Since video language models mostly originate from large language models~\cite{touvron2023llama,jiang2023mistral,he2025rodimus} with the help of visual instruction tuning~\cite{liu2024visual}, video language models can only resort to the language prior~\cite{goyal2017making,lin2023visualgptscore} when faced with knowledge incompleteness. This tendency brings spurious hallucination and misalignment with the visual facts~\cite{li2023evaluating, gunjal2024detecting} to the answers.
    
    The mainstream of video language models follows a single-video reasoning pipeline (shown in Fig.~\ref{fig:head_1}(a)). Motivated by the research in multi-data collaboration~\cite{wang2023seggpt, zhang2024makes, wu2020dynamic} and retrieval-augmented generation~\cite{lewis2020retrieval,edge2024local,guo2024lightrag}, a promising idea to address this limitation is integrating knowledge from multiple related videos. We illustrate some video question-answering examples in Fig.~\ref{fig:single_vs_multiple}. As the examples demonstrate, single-video reasoning suffers from incomplete and redundant information from an individual video. 
    These challenges prevent video language models from grasping valuable knowledge and lead to hallucinations and failures in answering the given questions. Introducing multiple highly related videos within the same domain can compensate for the missing information and enable the video language models to conclude and provide reliable answers.
    
    Despite the apparent advantages, the incremental information from multiple videos does not necessarily lead to better reasoning in real cases. As illustrated in Fig.~\ref{fig:head_1}(b), A direct multi-video joint strategy is to concatenate the visual features of multiple videos in the prompt before passing them to the video language models, which can bring an overwhelming number of tokens to video language models~\cite{li2023videochat,lin2024video,maaz2023video}. When exposed to long context, empirical evidence~\cite {hsieh2024ruler,liu2024lost} proves that LLMs are inclined to concentrate only on specific and limited segments of the input, neglecting potentially critical information. Moreover, the high dimensionality, redundancy, and obfuscation of video content contrast sharply with the structured, clear, and rule-based nature of language. This temporal visual complexity makes the synthesis and refinement of multi-video knowledge particularly challenging.
    
    In this paper, we present a structured multi-video collaborative reasoning framework. As illustrated in Fig.~\ref{fig:head_1}(c), our framework achieves the alignment and completion of multi-video information at the graph-structured representation level. Under this framework, to address the redundancy and complexity of video information, we first design the Video Structuring Module (VSM). 
    This module analyzes key targets and extracts their spatio-temporal relationships step by step, thereby constructing a data-efficient structured video representation. 
    After obtaining the structured representation, we propose the Graph Fusion Module (GFM) to yield LLM-friendly graph tokens. GFM integrates the structural information into node features through the graph attention network (GAT)~\cite{veličković2018graph} and then fuses the multi-video knowledge based on Cross-Graph Attention (CGA). Finally, we arrange the fused graph tokens, target video tokens, and text tokens with an elaborately designed prompt to augment VLMs with the multi-video structured knowledge. Overall, our contributions are threefold: 
    \begin{itemize}
        \item We explore a feasible structured reasoning framework for multi-video collaboration on video language models.
        \item Under this framework, we develop the Video Structuring Module to obtain the data-efficient spatio-temporal graph for video information. Then, we further design the Graph Fusion Module to fuse the spatio-temporal relations and cross-video knowledge into the graph tokens.  %
        \item Extensive experiments show that our framework can boost the reliability and accuracy through multi-video structuring and collaboration.
    \end{itemize}

\begin{figure*}
    \centering
    \includegraphics[width=\linewidth]{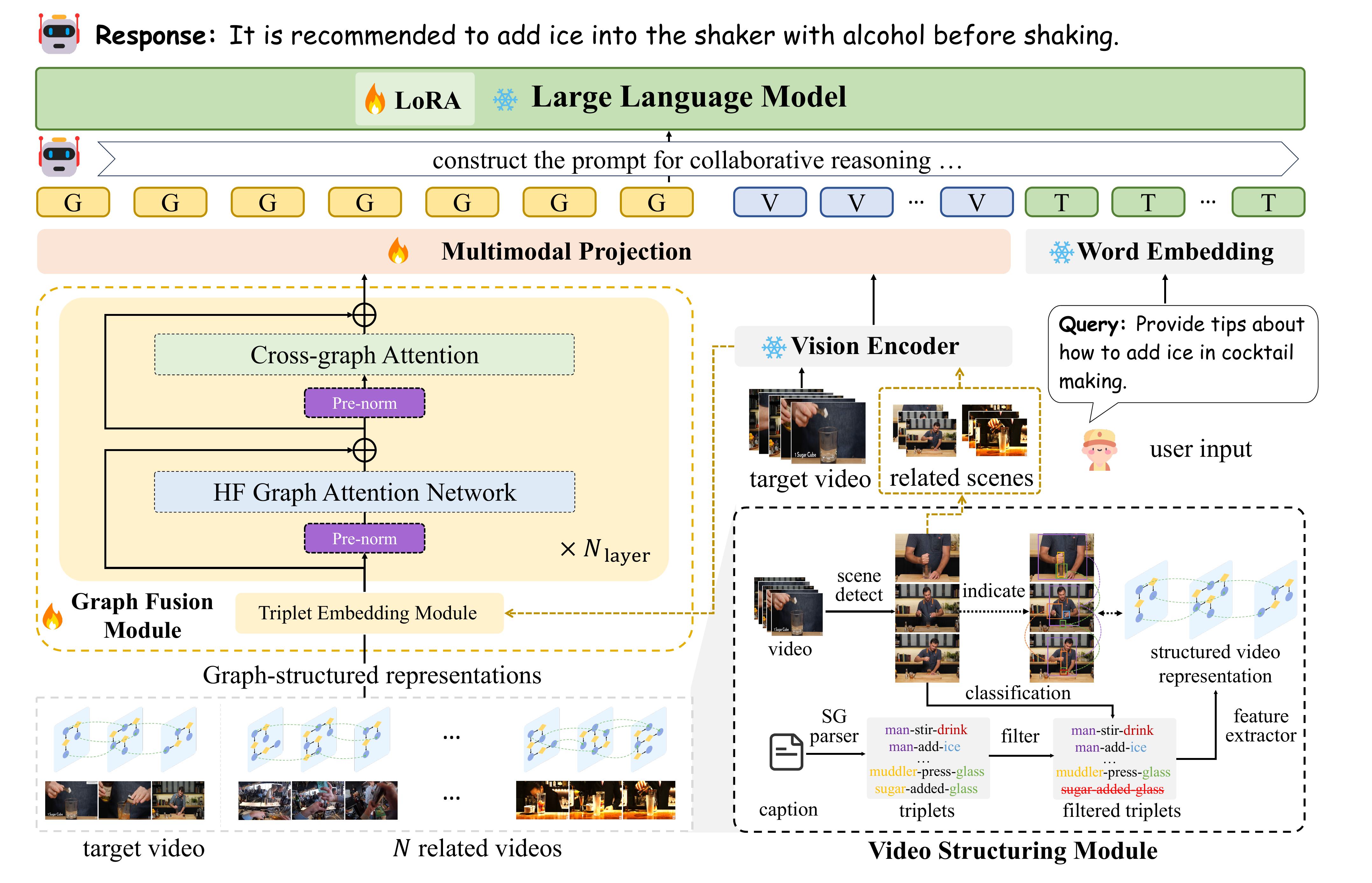}
    \caption{\textbf{Multi-video collaborative reasoning framework.} Together with the target video, $N$ related videos are retrieved to facilitate the reasoning process. First, we design the Video Structuring Module to obtain the structured video representation. Then, the Graph Fusion Module fuses the structure information and the related videos' information to get the video graph tokens. Finally, according to the designed prompts, the graph tokens, visual tokens, and text tokens are arranged as input to the large language model for question answering.}
    \label{fig:framework}
\end{figure*}
    
\section{Related Work}
\subsection{Vision Language Models}
    In the field of visual understanding, significant advancements have been achieved along with the development of large language models. By learning joint representations of vision and language through feature alignment and visual instruction tuning~\cite{liu2024visual}, vision-language models leverage the generalist capabilities that emerge from the large-scale training of large language models, enabling understanding of images and videos in open-world settings. Among them, Video-LLaMA~\cite{zhang2023video} and BLIP-2~\cite{li2023blip} adopt the Q-former to extract valuable visual information into compact and LLM-friendly visual tokens for alignment. On the contrary, Flamingo~\cite{alayrac2022flamingo} and BLIP-3~\cite{xue2024xgen} utilize the scalable perceiver resampler to obtain learnable visual tokens. LLaVA~\cite{liu2024visual}, Video-LLaVA~\cite{lin2024video}, LLaVA-OneVision~\cite{li2024llava}, Qwen2/2.5-VL~\cite{wang2024qwen2, bai2025qwen2} use an MLP-based projector to map the visual and textual inputs into a shared feature space. Besides, due to the complicated information contained in image and video input, many studies~\cite{li2023blip,li2024llama,cheng2024videollama,liu2025nvila} aim to reduce the number of visual tokens for alleviating the computational and understanding burden on LLMs. While current methods focus on single-video reasoning, we try to explore a structured video reasoning framework to effectively address the challenges of redundancy and integration brought by multi-video inputs. 
    
\subsection{Multi-Data Collaboration}
    Although most deep learning task follows a single data processing pipeline, multi-data collaboration provides a promising direction to improve performance with the inner correspondence among multiple samples. We can roughly divide current works into two streams, which are content-related and task-related collaborations. Content-related collaboration helps models to focus on vital content through comparison among several related data. For example, the co-segmentation task~\cite{zhang2024makes,wang2023seggpt} explores ways to segment the same object in different scenes by concluding and sharing the same object-related features. Few-shot image classification~\cite{hou2019cross}, action recognition~\cite{wang2022hybrid,cao2020few}, and fine-grained classification~\cite{luo2019cross} methods focus on the key differences for accurate classification through multi-data comparison. Retrieval-Augmented Generation~\cite{lewis2020retrieval,edge2024local,guo2024lightrag} is also a promising content-related collaboration that provides the LLMs with supporting information from retrieved related data. In task-related collaboration, models learn the way to complete the task by observing multiple samples performing the same task. For example, multi-video summarization~\cite{wu2019mvsgcn,wu2020dynamic,panda2017diversity} researches aim to generate a summary from a collection of videos through multi-data complementary and refinement. In-Context Learning~\cite{wang2023images,bar2022visual} uses examples with task guidance and answering to demonstrate to LLMs how to fulfill the tasks. 
    For video language models, the current multi-video collaboration strategy directly concatenates multiple inputs, leading to burdensome redundancy and difficulties in collaboration. Building on these advancements, we introduce the multi-video reasoning task and achieve the compensation and refinement of key spatio-temporal information through multi-video collaboration.

\section{Method}
\subsection{Overview}
\subsubsection{The setting of multi-video reasoning}
    In this paper, we explore the potential of utilizing multi-video information for video understanding. Specifically, we follow a multi-video setting, where a target video $V_0$ is accompanied by $N$ related videos $\{V_1, V_2, \dots, V_N\}$. To retrieve related videos, their feature vectors are constructed in advance for efficient video retrieval. Different ways of video vectorization are discussed in Sec.~\ref{sec:discussion}. Finally, the methods are required to answer the question about the target video with the help of $N$ retrieved videos.

\subsubsection{Framework overview}
    Our multi-video collaborative reasoning framework is illustrated in Fig.~\ref{fig:framework}. 
    First, the Video Structuring Module is introduced in Sec.~\ref{sec:represent} to obtain the structured video representation.
    Based on the obtained video structure, we design the Graph Fusion Module in Sec.~\ref{sec:graph_fusion} to fuse the structured video representation and convert useful related videos' information to the resulting graph tokens.
    Finally, we arrange all the graph, visual, and text tokens according to our designed multi-video reasoning prompt and send them to the large language model.

\begin{figure*}[t]
    \centering
    \includegraphics[width=0.9\linewidth]{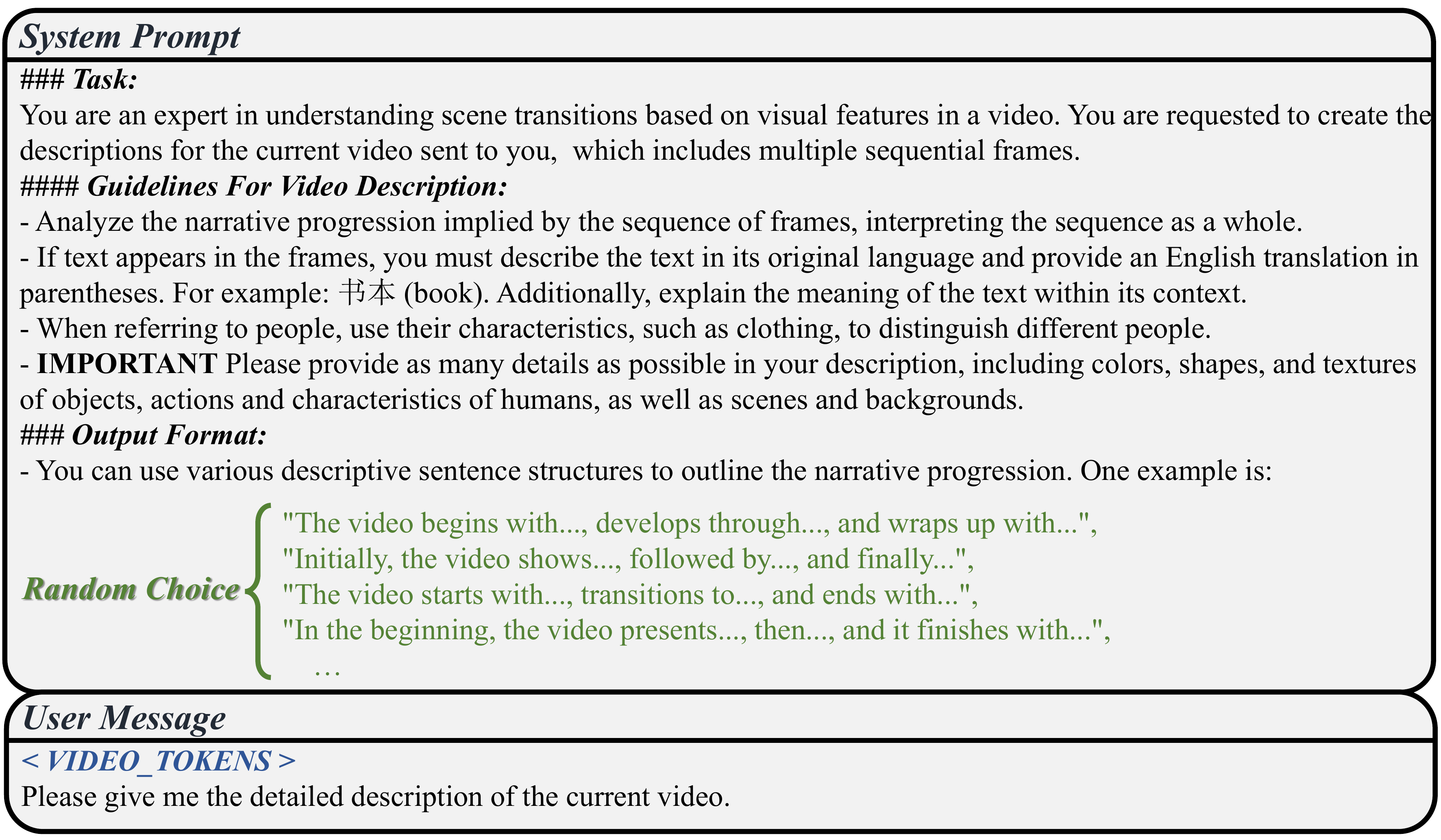}
    \caption{\textbf{Video captioning prompts.} We refer to the design outlined in \cite{zhang2024videoinstructiontuningsynthetic} to create the prompts used to extract captions from videos. The prompts are divided into two parts: the system prompt and the user message. In the system prompt, we define the task of video captioning and provide corresponding guidelines along with a standardized output format. For the output format, the program randomly selects contents in green font as the normalized format for reference during each process of captioning. For the user message, we utilize $<$\textbf{\textit{VIDEO\_TOKENS}}$>$ as the video tokens, and we provide a concise instruction to the model, then generate a detailed description for the video.}
    \label{fig:caption_prompt}
\end{figure*}

\begin{figure*}
    \centering
    \includegraphics[width=0.9\linewidth]{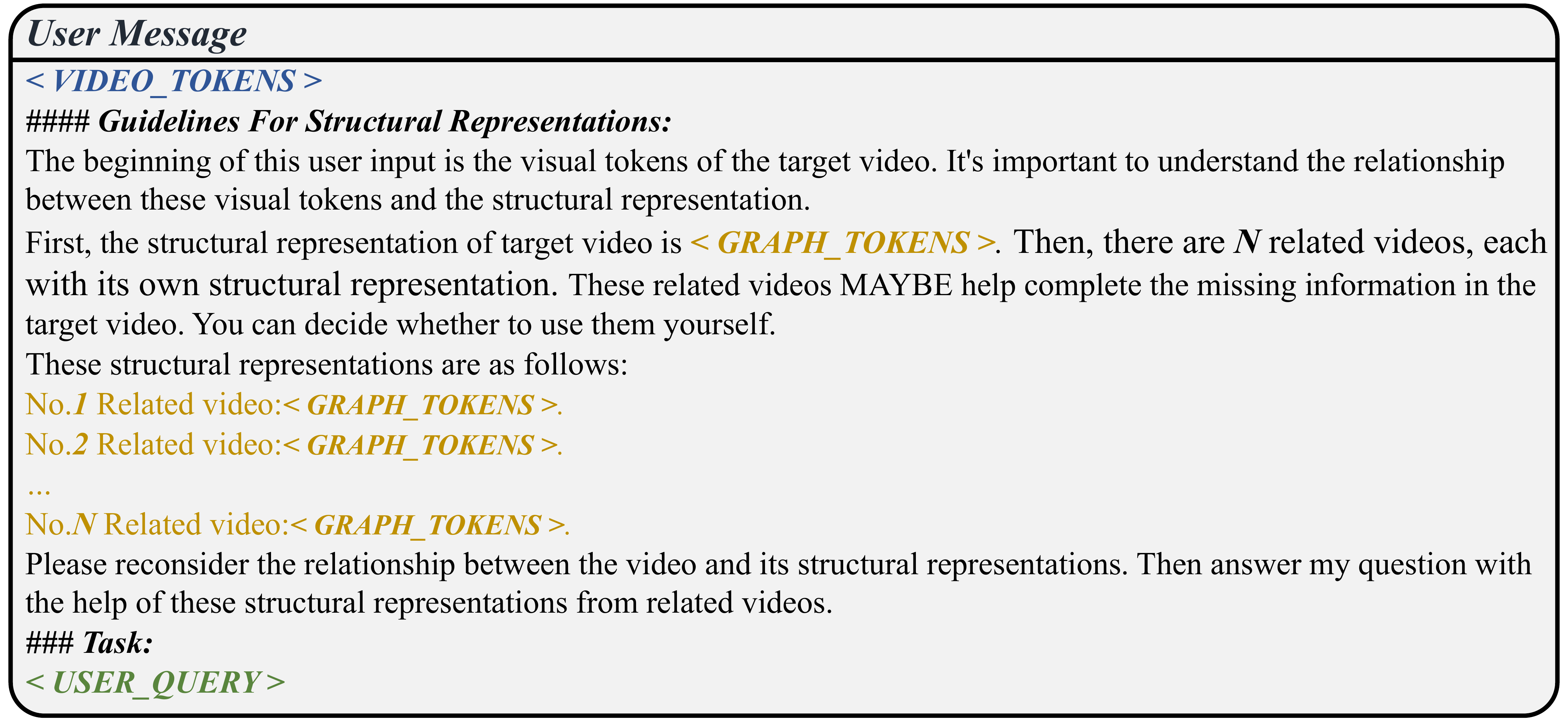}
    \caption{\textbf{Structured multi-video prompts.} We properly integrate the multi-modal tokens, together with the prompt guidance, to form an LLM-friendly input.}
    \label{fig:prompts}
\end{figure*}

\subsection{Video Structuring Module}
\label{sec:represent}
    The efficient structured video representation paves the way for the following integration of multi-video knowledge. Given the video and its paired dense caption, the process of our Video Structuring Module (VSM) is depicted on the bottom right of Fig.~\ref{fig:framework}, which can be specified as follows.

    \noindent\textbf{Step 1: Scene detection.} To reduce temporal redundancy within the video, we employ a lightweight, content-based scene detector, \textit{Autoshot}~\cite{zhu2023autoshot}, to segment the video into distinct scenes. From each detected scene, we extract its middle frame as a keyframe, which will serve as input for the subsequent video structuring process. Let the $M$ keyframes of video $V_N$ be denoted as $\mathcal{F}_N = \{F_1, F_2, \dots, F_M\}$.

    \noindent\textbf{Step 2: Dense video captioning.} To prepare for the latter structuring pipeline, we need to extract the detailed and fine-grained textual concept. 
    For this purpose, we leverage a video large language model to generate comprehensive descriptions of the input videos using the designed prompt, as detailed in Fig.~\ref{fig:caption_prompt}.

    \noindent\textbf{Step 3: Textual scene graph parsing.} We then extract the textual scene graph $\mathcal{G}^{\text{Text}}$ from the dense video caption using the \textit{SceneGraphParser}~\cite{wu2019unified}, replacing its large language model with Qwen3-30-A3B~\cite{yang2025qwen3}. 
    The textual scene graph comprises several triplets $\tau_i = \{s_i, p_i, o_i\}$, where each triplet $\tau_i$ represents the $i$-th interaction or event within the video. Here, $s_i$, $p_i$, and $o_i$ denote the subject, predicate, and object, respectively. Each triplet is formatted as \textlangle subject - predicate - object\textrangle, serving as a foundational representation of the relationships and dynamics captured in the video.
    
    \noindent\textbf{Step 4: Graph information filtering.} 
    To improve data quality, we apply a proactive filtering mechanism to eliminate irrelevant or redundant triplets. Specifically, we use an image-level classifier to verify the presence of relevant objects or subjects from the triplets within the scene. This is achieved by formulating a simple binary classification task with tailored prompts and utilizing SigLIP~\cite{zhai2023sigmoid} to perform the classification. For instance, we design prompts such as, ``\textit{The object related to \{object/subject\} is in the image.}'' for positive samples and, ``\textit{The object related to \{object/subject\} is not in the image.}'' for negative samples. Based on the classification results, we determine whether to retain or discard the object-subject pair.
    In cases where either the object or subject exists independently in the image, we construct triplets in the format $\{s_i, *, s_i\}$ or $\{o_i, *, o_i\}$, to establish self-connections between nodes in the next step. Conversely, if neither the object nor the subject is present simultaneously, the corresponding triple will be discarded. This process yields the filtered triplets, denoted as $\hat{\mathcal{G}}^{\text{Text}}$.
    
    \noindent\textbf{Step 5: Video graph establishment.} 
    Built upon the filtered textual scene graph $\hat{\mathcal{G}}^{\text{Text}}$ and the keyframes $\mathcal{F}_{\{0,\dots,N\}}$ corresponding to each triplet, we establish the graph-based structured video representation for both the target video and related videos. Specifically, the graph consists of the \textit{nodes}, \textit{intra-frame edges}, and \textit{inter-frame edges}. The \textit{node} represents the features of the object or subject in the video. For node-level features, we first leverage Qwen3-Embedding~\cite{zhang2025qwen3} to extract the text features $\mathbf{T}$ from $\hat{\mathcal{G}}^{\text{Text}}$, transforming the text into feature representation for each object and subject, then, we introduce Pooling Attention to extract effective visual features based on text features and the keyframes $\mathcal{F}_{\{0,\dots,N\}}$ and perform adaptive weighted fusion with original text features to obtain a more robust node-level feature representation. Further details regarding this process will be provided in Sec.~\ref{sec:graph_fusion}.
    The \textit{intra-frame edge} is represented by the predicate of each triplet, representing spatial and interacting relationships between objects and objects within one frame. Meanwhile, the \textit{inter-frame edge} links objects across different frames that share the same subjects and objects, thereby modeling their temporal relationships. Through the above steps, we can establish the graph-based structured video representation of both the target video and its related videos for further collaboration.

\subsection{Graph Fusion Module}
\label{sec:graph_fusion}
    The Graph Fusion Module (GFM) consists of a Triplet Embedding Module (TEM) and a multi-layer stacked architecture designed for graph information processing. Each layer of the architecture integrates two essential components: the Hierarchical Frame Graph Attention Network (HF-GAT) and the Cross-Graph Attention (CGA) mechanism.

    First, within the TEM, we introduce class embedding (CE) to enhance the ability of the GFM to distinguish between the graph of the target video and the graphs of related videos. The CE is defined as follows:
    \begin{equation}
    \label{eq:class_pe}
    \begin{aligned}
        \mathrm{CE}_{tar} &= \sigma(\boldsymbol{\alpha}),
        \\
        \mathrm{CE}_{rel} &= 1 - \sigma(\boldsymbol{\alpha}),
    \end{aligned}
    \end{equation}
    where $\boldsymbol{\alpha} \in \mathbb{R}^{d}$ represents a learnable parameter shared across frames, and $\sigma$ denotes the sigmoid function.
    The computed class embeddings are then directly applied to the inputs of the GFM, with $\mathrm{CE}_{tar}$ used for the text features $\mathbf{T}_{tar}$ from the target video and $\mathrm{CE}_{rel}$ for other text features $\mathbf{T}_{rel}$ from the related videos. This integration process is formulated as follows:
    \begin{equation}
    \label{eq:add_class_pe}
    \begin{aligned}
          \mathbf{T}_{tar} &= \mathbf{T}_{tar} + \mathrm{CE}_{tar},
          \\
          \mathbf{T}_{rel} &= \mathbf{T}_{rel} + \mathrm{CE}_{rel},
          \\
          \mathbf{T} &= [\mathbf{T}_{tar}, \mathbf{T}_{rel}],
    \end{aligned}
    \end{equation}
    where $[,]$ indicates the concatenation operation applied to the triplets from the target video and related videos.
    By leveraging CE, the GFM can implicitly learn to distinguish between the target video and the related videos. 
    
    Additionally, to effectively incorporate the visual information from the keyframes corresponding to the triplets, we integrate Pooling Attention, as defined in Eq.~\eqref{eq:pooling_attention}, within the TEM, enabling direct aggregation of visual features from the keyframes guided by text features.
    \begin{equation}
    \label{eq:pooling_attention}
    \begin{aligned}
        \mathbf{Q}&=\mathbf{T}\mathbf{W}_Q \in \mathbb{R}^{1\times d},
        \\
        \mathbf{K}&=\mathbf{I}\mathbf{W}_K \in \mathbb{R}^{(H_p\times W_p)\times d},
        \\
        \mathbf{V}&=\mathbf{I}\mathbf{W}_V \in \mathbb{R}^{(H_p\times W_p)\times d},
        \\
        \mathbf{\tilde{I}} &=  \mathrm{softmax}(\mathbf{Q} \mathbf{K}^\top / \sqrt{d}) \mathbf{V} \in \mathbb{R}^{1 \times d}, 
    \end{aligned}
    \end{equation}
    where $\mathbf{W}_Q \in \mathbb{R}^{d \times d},\mathbf{W}_K \in \mathbb{R}^{d \times d},\mathbf{W}_V \in \mathbb{R}^{d \times d}$ are learnable weight matrices associated with the queries $\mathbf{Q}$, the keys $\mathbf{K}$, and values $\mathbf{V}$ of attention mechanism~\cite{vaswani2017attention}, respectively, and the $\mathbf{I} \in \mathbb{R}^{(H_p\times W_p)\times d}$ represents the visual features extracted from the vision encoder of VLMs, which utilize multiple tokens to represent a keyframe. $(H_p\times W_p)$ denotes the length of visual features after extraction from the vision encoder.
    By applying Pooling Attention, we aggregate the visual features in a manner guided by the text features, resulting in a more robust feature representation.
    Subsequently, we fuse the original text features $\mathbf{T}$, extracted from Qwen3-Embedding~\cite{zhang2025qwen3} and the pooled visual features $\mathbf{\tilde{I}}$ using adaptive weights $\boldsymbol{\beta} \in \mathbb{R}^{d}$, as defined by:
    \begin{equation}
    \label{eq:add_pooling_attention}
    \mathbf{\hat{T}} = \sigma(\boldsymbol{\beta}) \odot \mathbf{T} + (1 - \sigma(\boldsymbol{\beta})) \odot \mathbf{\tilde{I}},
    \end{equation}
    where $\odot$ denotes the Hadamard product (element-wise multiplication). This fusion operation adaptively balances the contributions of text and visual features, yielding a final robust representation $\mathbf{\hat{T}}$.
    
    Then, we input the processed triplet features $\mathbf{\hat{T}}$ into a multi-layer architecture to process the graph information. The features are first passed into HF-GAT, which is tailored to fuse graph-based structured data within a single video. While traditional Graph Attention Networks (GATs)~\cite{veličković2018graph} are primarily used for tasks such as node classification and relation prediction, where relationships between nodes are explicitly defined. In contrast, for both inter-frame and intra-frame contexts, these relationships are often implicit or absent. 
    To address this challenge, as introduced in Sec.~\ref{sec:represent}, we first convert the original visual modality data into a graph-based structured representation with the VSM. In the constructed graph, nodes represent features of subjects or objects, which are initially extracted from Sec.~\ref{sec:represent} and subsequently processed by the TEM.
    For intra-frame edges, we utilize relationships based on triplets, where a directional link is formed from $s_i$ to $o_i$. For inter-frame edges, we leverage the filtered results $\hat{\mathcal{G}}^{\text{Text}}$ obtained in Step 4 (Sec.~\ref{sec:represent}) to link $s_i^{t-1}$ or $o_i^{t-1}$ from the previous frame to $s_i^{t}$ or $o_i^{t}$ in the current frame. Additionally, we introduce bidirectional links for inter-frame connections, as it enhances the system's ability to comprehend video content~\cite{simonyan2014two}.

    Once the structured features from individual videos are extracted with HF-GAT, the next step is to identify and fuse the most relevant information between videos.
    To achieve this, we introduce the Cross-Graph Attention mechanism, implemented through a self-attention mechanism with custom position IDs.
    To facilitate multi-video collaborative reasoning via triplet features, we identify three key principles:
    1) Within a single triplet, the relationship between the subject and object is non-exchangeable.
    2) Within a single video, the positional relationship between triplets is unordered.
    3) Across multiple videos retrieved via relevance ranking, the order of triplets between videos is non-exchangeable.
    For principles 1) and 2), the structured video representation from the HF-GAT inherently captures the position encoding within the individual video, as the HF-GAT aggregates and passes information between representations based on the connection relationship of the graph of the corresponding video, thereby implicitly encoding position information in the representation. Consequently, our primary focus is on addressing principle 3) while ensuring compliance with principles 1) and 2).
    To handle principle 3), we assign consistent position IDs within each video and dynamically adjust position IDs across retrieved videos based on their relevance ranking. For instance, triplet features from the target video are always assigned a position ID of 0, whereas position IDs for related videos are determined dynamically depending on their rank derived from retrieval relevance. These position IDs are then integrated through RoPE~\cite{su2024roformer} to encode positional information effectively in the Cross-Graph Attention mechanism.
    
    Moreover, we also apply residual connections and pre-normalization for both the HF-GAT and CGA components. Specifically, we apply the LayerNorm~\cite{ba2016layer}, commonly used in Vision Transformers (ViT)~\cite{zhai2023sigmoid}, for pre-normalization. Notably, we exclude the Feed-Forward Network (FFN) in layers to preserve the invariance of aligned visual features from the vision encoder, minimizing excessive feature shifting and ensuring linearity between inputs and outputs of GFM~\cite{gao2024linvt}.
    The graph-based structured video representations of the videos are then processed through the GFM to construct graph tokens, each corresponding to the node features of subjects or objects after being fused with the structured information.

\subsection{Structured Multi-Video Prompt}
    Upon obtaining the fused multi-video graph tokens, we need to integrate the graph, video, and text tokens together to create an LLM-friendly input. Therefore, we propose the structured multi-video prompt, as is shown in Fig.~\ref{fig:prompts}. 
    Our prompt originates from previous video language models' prompt design~\cite{lin2024video}. For the target video, we maintain the video tokens $<$\textbf{\textit{VIDEO\_TOKENS}}$>$ of the target video to preserve fine-grained and detailed visual information. We also append its graph-based structured data $<$\textbf{\textit{GRAPH\_TOKENS}}$>$ to indicate the key object and spatio-temporal relationships. For $N$ related videos, we only maintain the concise and data-efficient graph-based structured data $<$\textbf{\textit{GRAPH\_TOKENS}}$>$. In this context, we further indicate the relationships between the target video and related videos and how the LLMs can utilize these related multi-video structured data. 
    By structuring the prompt in this manner, we enable the video language model to leverage multi-video information in an effective way, thereby enhancing the model's ability to reason about and answer queries related to the video content.

\section{Experiments}
\subsection{Experiment Setup}
\label{sec:implementation}
    \noindent\textbf{Dataset descriptions.}
    During the training phase, we construct a dataset containing structured video information for GFM training, based on the LLaVA-Video-178K dataset~\cite{zhang2024videoinstructiontuningsynthetic}. Specifically, we perform step-by-step preprocessing of the dataset as outlined in the Sec.~\ref{sec:represent}. Notably, for video data in LLaVA-Video-178K that already includes captions, we retain the original captions without additional modifications to streamline the preprocessing workflow. Regarding video vectorization and retrieval mechanisms for related videos, we utilize Qwen3-Embedding-8B~\cite{zhang2025qwen3} to extract query embeddings (for retrieval) and document embeddings (for storage) from video captions. This approach is widely used in retrieval tasks~\cite{zhang2025qwen3}. Specifically, document embeddings are generated by directly inputting the captions, while query embeddings are generated with the designed prompt to prepare the input for the embedding model, as shown below:
    \begin{lstlisting}[xleftmargin=0pt, xrightmargin=0pt]
Instruct: This is the caption of a video. 
Please provide a search query to retrieve the caption representation of the other most relevant videos. \n
Query: {caption}.
    \end{lstlisting}

    Finally, we construct a training dataset comprising approximately 87K samples. 
    While relatively small compared to datasets commonly used for training other VLMs (e.g., 87K vs. 9.36M~\cite{li2024llava}), our method achieves effective performance improvements, as shown in Tab.~\ref{tab:perform}. 
    This result highlights the ability of our method to integrate seamlessly into existing model frameworks and deliver performance gains through simple and efficient training on a compact dataset.

    For evaluation, we test our approach on different video question-answering benchmarks, including 
    ActivityNet-QA~\cite{yu2019activitynet}, NExT-QA~\cite{xiao2021next}, EgoSchema~\cite{mangalam2023egoschema}, and Video-MME~\cite{fu2025video}. 
    These benchmarks cover both short and long video understanding tasks, providing a comprehensive evaluation of our method.
    
    Additionally, to enhance experimental efficiency for the ablation study, we utilize approximately 10\% of the original training dataset for training. Similarly, during evaluation, we select subsets of the NExT-QA and EgoSchema datasets, each containing around 0.5K samples.

    \noindent\textbf{Implementation details.}
    Our proposed structured multi-video collaboration framework is adaptive to general video language models. 
    To verify the effectiveness of our proposed method, we conduct experiments with various models of different parameter scales on A6000 48GB GPUs, including LLaVA-OneVision-0.5B~\cite{li2024llava} and LLaVA-Video-7B~\cite{zhang2024videoinstructiontuningsynthetic}. 
    For the Graph Fusion Module (GFM), the hidden state size is configured to match the output dimension of the corresponding vision encoder. We initialize the video language model (VLM) with pre-trained weights and further enhance its ability to effectively comprehend graph-based video representations by training it on our constructed dataset.
    To optimize our model efficiently, we adopt a standard two-stage training strategy~\cite{lin2024video,gao2024linvt}. During the first stage, we freeze the vision encoder, projector, and language model, focusing on training the GFM to align the inputs to the language model. In the second stage, we unfreeze the projector and language model while keeping the vision encoder frozen, and apply LoRA~\cite{hu2022lora} to the language model, then fine-tune the projector, GFM, and language model simultaneously.
    The detailed training recipe and hyperparameter configurations are provided in Tab.~\ref{tab:train_recipe}.
    

\begin{flushleft}
\centering
\begin{minipage}{0.48\textwidth}
\captionof{table}{\textbf{The training recipe for VLM in our experiments.}}
\label{tab:train_recipe}
\centering
\setlength{\belowcaptionskip}{0.1cm}
\renewcommand\arraystretch{1.1}
\resizebox{\linewidth}{!}{
\begin{tabular}{@{}l|c|c@{}}
\toprule[1pt]
                          & \textbf{Stage-1} & \textbf{Stage-2}           \\ \midrule
Trainable                 & GFM     & GFM, Projector, LLM \\ \midrule
Batch size                & 128     & 64                \\ 
Optimizer & AdamW & AdamW \\
Warmup ratio & 0.03 & 0.03 \\
Learning rate schedule & Cosine decay & Cosine decay \\
\midrule
LR: $\boldsymbol{\phi}_{\text{GFM}}$   & 1e-3    & 1e-4              \\
LR: $\boldsymbol{\phi}_{\text{Proj.}}$ & -       & 1e-5              \\
LR: $\boldsymbol{\phi}_{\text{LLM}}$   & -       & 1e-5              \\ \bottomrule[1pt]
\end{tabular}
}

\end{minipage}
\end{flushleft}

\begin{table*}
    \centering
    \setlength{\belowcaptionskip}{0.1cm}
    \renewcommand\arraystretch{1.1}
    \caption{\textbf{Video question answering performances on different large video-language models.} The \uwave{wavy lines} indicate the re-evaluated results.}
    \label{tab:perform}
    \resizebox{\textwidth}{!}{
    \begin{tabular}{@{}l|c|c|c|c|c|c|c@{}}
    \toprule[1pt]
    \textbf{Model} & \textbf{Params} & \textbf{Frames} & \textbf{ActivityNet-QA} & \textbf{NExT-QA} & \textbf{EgoSchema} & \textbf{Video-MME} & \textbf{Average} \\
    \scriptsize Task &  &  & \scriptsize Open-Ended & \scriptsize Multi-Choice & \scriptsize Multi-Choice & \scriptsize Multi-Choice & \scriptsize Acc. (\%)  \\
    \scriptsize Duration &  &  & \scriptsize Short & \scriptsize Short & \scriptsize Long & \scriptsize Long &  \\
    \midrule
    Video-LLaVA~\cite{lin2024video} & 7B & 8 & 45.30 & 62.60 & 38.40 & 40.40 & 46.68 \\ 
    LLaMA-VID~\cite{li2024llama} & 7B & 1fps & 47.40 & - & 38.50 & - & - \\ 
    PLLaVA~\cite{xu2024pllava} & 7B & 16 & 56.30 & 68.17 & 45.16 & 44.25 & 53.47 \\ 
    VideoChat2~\cite{li2024mvbench} & 7B & 16 & - & - & 54.40 & 47.90 & - \\ 
    LLaVA-NeXT-Video~\cite{zhang2024llavanextvideo} & 7B & 32 & 53.50 & - & 43.90 & 46.50 & - \\ 
    Qwen2-VL~\cite{wang2024qwen2} & 7B & 2fps & 57.40 & 77.20 & 66.70 & 63.30 & 66.15 \\ 
    Qwen2.5-VL~\cite{bai2025qwen2} & 3B & 2fps & - & - & 64.80 & 61.50 & - \\ 
    Qwen2.5-VL~\cite{bai2025qwen2} & 7B & 2fps & - & - & 65.00 & 65.10 & - \\ 
    VideoLLaMA2~\cite{cheng2024videollama} & 7B & 16 & 50.20 & 75.60 & - & 47.90 & - \\ 
    VideoLLaMA2.1~\cite{cheng2024videollama} & 7B & 16 & 53.00 & 75.60 & 53.10 & 54.90 & 59.15 \\ 
    VideoLLaMA3~\cite{zhang2025videollama} & 2B & 180 & 58.20 & 81.10 & 58.50 & 59.60 & 64.35 \\ 
    InternVL2~\cite{chen2024expanding} & 8B & 16 & - & - & 55.00 & 54.00 & - \\ 
    InternVL2.5~\cite{chen2024expanding} & 8B & 64 & 58.90 & 85.00 & 51.50 & 64.20 & 64.90 \\ 
    NVILA~\cite{liu2025nvila} & 8B & 256 & 60.90 & 82.20 & 54.30 & 64.20 & 65.40 \\ 
    \midrule
    LLaVA-OneVision~\cite{li2024llava} & 0.5B & 32 & \uwave{45.65} & 57.20 & 26.80 & \textbf{44.00} & 43.41 \\ 
    \quad \textbf{+Ours} & 0.5B & 32 & \textbf{46.46} & \textbf{58.7}1 & \textbf{28.38} & 43.74 & \textbf{44.32} \\
    \cdashline{1-8}[1pt/1pt]
    LLaVA-Video~\cite{zhang2024videoinstructiontuningsynthetic} & 7B & 64 & \uwave{60.55} & 83.20 & 57.30 & 63.30 & 66.09 \\ 
    \quad \textbf{+Ours} & 7B & 64 & \textbf{61.25} & \textbf{84.00} & \textbf{61.76} & \textbf{64.37} & \textbf{67.84} \\ 
    \bottomrule[1pt]
    \end{tabular}
    }
\end{table*}

\subsection{Video Question Answering}

    We evaluate advanced video language models on video question-answering tasks, including ActivityNet-QA~\cite{yu2019activitynet}, NExT-QA~\cite{xiao2021next}, EgoSchema~\cite{mangalam2023egoschema}, and Video-MME~\cite{fu2025video}, which collectively span a diverse range of video understanding tasks.
    Following the methodology proposed by Video-LLaVA~\cite{lin2024video}, we employ ChatGPT-Assistant to report the accuracy of open-ended answers for the ActivityNet-QA dataset. 
    However, since the gpt-3.5-turbo-0613 model used in ActivityNet-QA's original evaluation pipeline has been deprecated, for fair comparison, we opt to re-evaluate the results using Qwen3-235B-A22B~\cite{yang2025qwen3}, an open-source large language model. Being open-source, this model is easier to access and utilize compared to the GPT series. Moreover, it exhibits superior language capabilities over the gpt-3.5-turbo series~\cite{yang2025qwen3}, making it an ideal replacement for evaluating the correctness of open-ended answers.
    Thus, to ensure a fair and consistent comparison of baselines, we use Qwen3-235B-A22B to perform the re-evaluation of the ActivityNet-QA dataset for LLaVA-OneVision-0.5B and LLaVA-Video-7B.
    We further compare the performance of several advanced video language models, all of which perform video question answering conditioned on a single video. Unlike these traditional methods, our framework extends the reasoning capabilities by generating answers based on both the target video and multiple retrieved related videos.
    The experimental results, as presented in Tab.~\ref{tab:perform}, highlight the superiority of our method over the baselines LLaVA-OneVision-0.5B and LLaVA-Video-7B. By introducing the concept of multi-video collaborative reasoning, our approach enhances average accuracy across various tasks, including open-ended answering, multiple-choice question answering, and video comprehension tasks covering videos of diverse durations.
    These results demonstrate that our method efficiently trains on the compact dataset, integrates multi-video knowledge, and provides more reliable answers.

\begin{table*}[ht]
\begin{minipage}{0.48\textwidth}
\centering
\setlength{\belowcaptionskip}{0.1cm}
\renewcommand\arraystretch{1.1}
\caption{\textbf{Ablation study on video structuring and multi-video fusion components on NExT-QA}, conducted using the baseline model LLaVA-OneVision-0.5B~\cite{li2024llava}.}
\label{tab:diff_prompt-llava_ov_0b5}
\resizebox{\linewidth}{!}{
\begin{tabular}{@{}c|c|c|c@{}}
\toprule[1pt]
\textbf{Struct} & \textbf{Multi-video}   & \textbf{context $L$} & \textbf{NExT-QA} \\ \midrule
                & single video           & 6.5K                 & 61.4             \\
                & multi-video tokens (\textbf{32}) & 38K                  & \textbf{OOM}              \\
                & multi-video tokens (\textbf{8})  & 15K                  & 51.8             \\
                & multi-video captions   & 9.3K                 & 61.8             \\
                \cdashline{1-4}[1pt/1pt]
\checkmark      & single video           & 7.3K                 & 62.0             \\
\checkmark      & graph fusion module    & 7.5K                 & \textbf{65.2}    \\ \bottomrule[1pt]
\end{tabular}
}
\end{minipage}
\hfill
\begin{minipage}{0.48\textwidth}
\centering
\setlength{\belowcaptionskip}{0.1cm}
\renewcommand\arraystretch{1.1}
\caption{\textbf{Ablation study on video structuring and multi-video fusion components on NExT-QA}, conducted using the baseline model LLaVA-Video-7B~\cite{zhang2024videoinstructiontuningsynthetic}.}
\label{tab:diff_prompt-llava_video_7b}
\resizebox{\linewidth}{!}{
\begin{tabular}{@{}c|c|c|c@{}}
\toprule[1pt]
\textbf{Struct} & \textbf{Multi-video}   & \textbf{context $L$} & \textbf{NExT-QA} \\ \midrule
                & single video           & 13K                 & 79.8             \\
                & multi-video tokens (\textbf{64}) & 73K                  & \textbf{OOM}              \\
                & multi-video tokens (\textbf{8})  & 22K                  & 72.6             \\
                & multi-video captions   & 16K                 & 79.8             \\
                \cdashline{1-4}[1pt/1pt]
\checkmark      & single video           & 13.8K                 & 83.6             \\
\checkmark      & graph fusion module    & 14K                 & \textbf{84.2}    \\ \bottomrule[1pt]
\end{tabular}
}
\end{minipage}
\end{table*}

\begin{table*}[ht]
\begin{minipage}[t]{0.48\textwidth}
\centering
\setlength{\belowcaptionskip}{0.1cm}
\renewcommand\arraystretch{1.}
\caption{\textbf{Ablation on the design of GFM.} \textbf{PA} refers to Pooling Attention.}
\label{tab:ablation}
\resizebox{\linewidth}{!}{
\begin{tabular}{@{}cccc|cc@{}}
\toprule[1pt]
\textbf{HF-GAT} & \textbf{PA} & \textbf{CGA}        & \textbf{FFN} & \textbf{NExT-QA} & \textbf{EgoSchema}  \\ \midrule
                &             &            &              & 61.4   & 26.4 \\
\cdashline{1-6}[1pt/1pt]
\checkmark      &             &            &              & 64.2   & 28.0 \\
\checkmark      & \checkmark  &            &              & 64.4   & 28.2 \\
\checkmark      & \checkmark  & \checkmark &              & \textbf{65.0}   & \textbf{28.6} \\
\checkmark      & \checkmark  & \checkmark & \checkmark   & 64.4   & 27.6 \\ \bottomrule[1pt]
\end{tabular}
}
\end{minipage}
\hfill
\begin{minipage}[t]{0.48\textwidth}
\centering
\setlength{\belowcaptionskip}{0.1cm}
\renewcommand\arraystretch{1.2}
\caption{\textbf{Ablations on different video retrieval strategies.}}
\label{tab:retrieval}
\resizebox{\linewidth}{!}{
\begin{tabular}{@{}c|c|c@{}}
\toprule[1pt]
\textbf{Video Retrieval Strategy}       & \textbf{NExT-QA} & \textbf{EgoSchema} \\ \midrule
video vector-based retrieval   & 63.8    & 27.6      \\
restricted retrieval          & 63.6    & 27.6      \\
caption vector-based retrieval & \textbf{65.0}    & \textbf{28.6}      \\ \bottomrule[1pt]
\end{tabular}
}
\end{minipage}
\end{table*}

\begin{figure*}[t]
    \centering
    \begin{minipage}{0.45\textwidth}
        \centering
        \includegraphics[width=\textwidth]{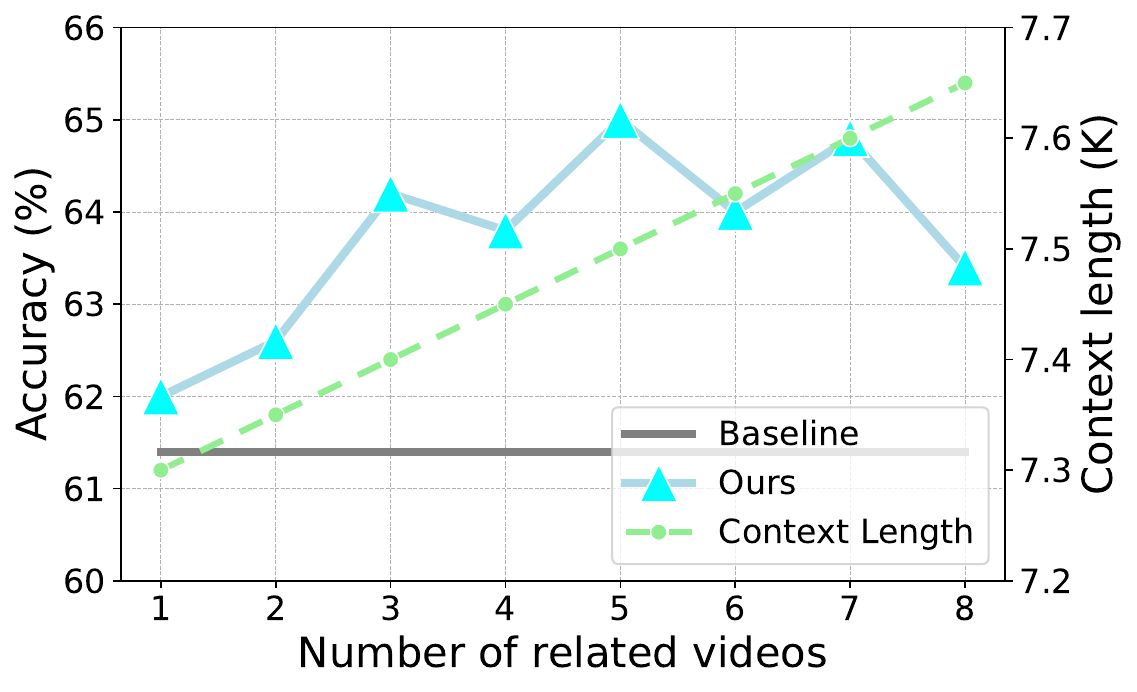}
        \caption*{(a) LLaVA-OneVision-0.5B.}
    \end{minipage}
    \hspace{10pt}
    \begin{minipage}{0.45\textwidth}
        \centering
        \includegraphics[width=\textwidth]{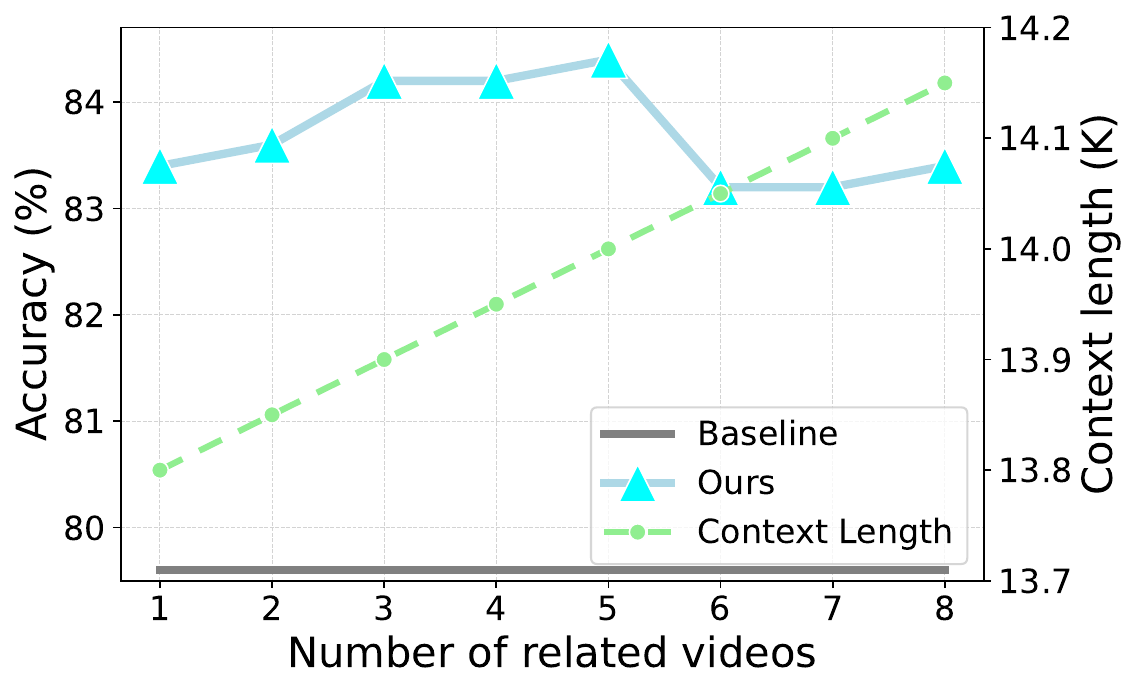}
        \caption*{(b) LLaVA-Video-7B.}
    \end{minipage}
    \caption{\textbf{Comparative analysis} of accuracy (\%) and context length (K) for NExT-QA across different models under varying numbers of related videos.}
    \label{fig:video_num}
\end{figure*}

\subsection{Ablation study}
\label{sec:abl}
    \noindent\textbf{Ablation study on the proposed components.}
    Our framework consists of two components: video structuring and multi-video collaboration. 
    The experimental results are shown in Tab.~\ref{tab:diff_prompt-llava_ov_0b5} for LLaVA-OneVision-0.5B~\cite{li2024llava} and Tab.~\ref{tab:diff_prompt-llava_video_7b} for LLaVA-Video-7B~\cite{zhang2024videoinstructiontuningsynthetic}. 
    First, we evaluate our framework by disabling the video structuring module and testing common multi-video fusion strategies. The ``multi-video tokens'' strategy involves concatenating all video tokens as input to the context, whereas the ``multi-video captions'' strategy sends the captions of all videos into the context.
    Initially, we use the default number of frames preset in the respective models for related videos, but this causes out-of-memory (OOM) issues during inference. This highlights that directly utilizing the ``multi-video tokens'' approach is impractical for real-world inference scenarios. We address this by reducing the number of frames for related videos to 8, while maintaining the default number of frames for the target video. Despite this adjustment, the ``multi-video tokens'' strategy introduces an excessive number of tokens into the context, challenging the model's understanding and leading to noticeable performance degradation (-9.6\% for LLaVA-OneVision-0.5B and -7.2\% for LLaVA-Video-7B). Conversely, the ``multi-video captions'' approach demonstrates slight performance improvements (+0.4\% for LLaVA-OneVision-0.5B), but performance remains unchanged for LLaVA-Video-7B.
    Enabling the Video Structuring Module provides a structured and clearer video representation, which helps large language models (LLMs) achieve better content understanding. This results in consistent performance improvements across models (+0.6\% for LLaVA-OneVision-0.5B and +3.79\% for LLaVA-Video-7B). Lastly, applying the multi-video graph fusion strategy allows the framework to extract valuable information from related videos effectively. This achieves substantial performance improvements (+3.8\% for LLaVA-OneVision-0.5B and +4.4\% for LLaVA-Video-7B) with minimal overhead, requiring only 0.2K additional tokens over single-video processing while leveraging the Video Structuring Module for enhanced reasoning.

    \noindent\textbf{Ablation on Graph Fusion Module design.}
    We further conduct an ablation study on the design of our Graph Fusion Module, which comprises three components: HF-GAT, CGA, and the Pooling Attention in the TEM. 
    To expedite both the training and evaluation, we perform these studies using the LLaVA-OneVision-0.5B~\cite{li2024llava} model, utilizing subsets of the original training and evaluation datasets to validate the effectiveness of the proposed components.
    Since we have discussed the usage of FFN in Sec.~\ref{sec:graph_fusion}, we also take it into consideration and conduct the ablation study on these three components.  The results of the ablation experiments are presented in Tab.~\ref{tab:ablation}. In the first row, the graph structure feature tokens are directly sent to the multimodal projection layer to get the graph tokens. In the second row, we incorporate HF-GAT to propagate structural information to each token, which increases performance by 2.8\% on the NExT-QA dataset and 1.6\% on the EgoSchema dataset.
    We also integrate Pooling Attention into TEM to embed structural information from specific scenes. The results demonstrate that incorporating this enriched information leads to a slight improvement compared to the original textual structural information (+3\% on the NExT-QA, +1.8\% on the EgoSchema). Furthermore, with the inclusion of CGA, multi-video knowledge is fused into graph tokens, resulting in further performance gains of 3.6\% on the NExT-QA dataset and 2.2\% on the EgoSchema dataset.
    Interestingly, the adoption of FFN does not improve reasoning accuracy, further supporting the conclusions drawn in Sec.~\ref{sec:graph_fusion}.

\begin{figure*}[t]
    \centering
    \begin{minipage}{0.43\textwidth}
        \centering
        \includegraphics[width=\textwidth]{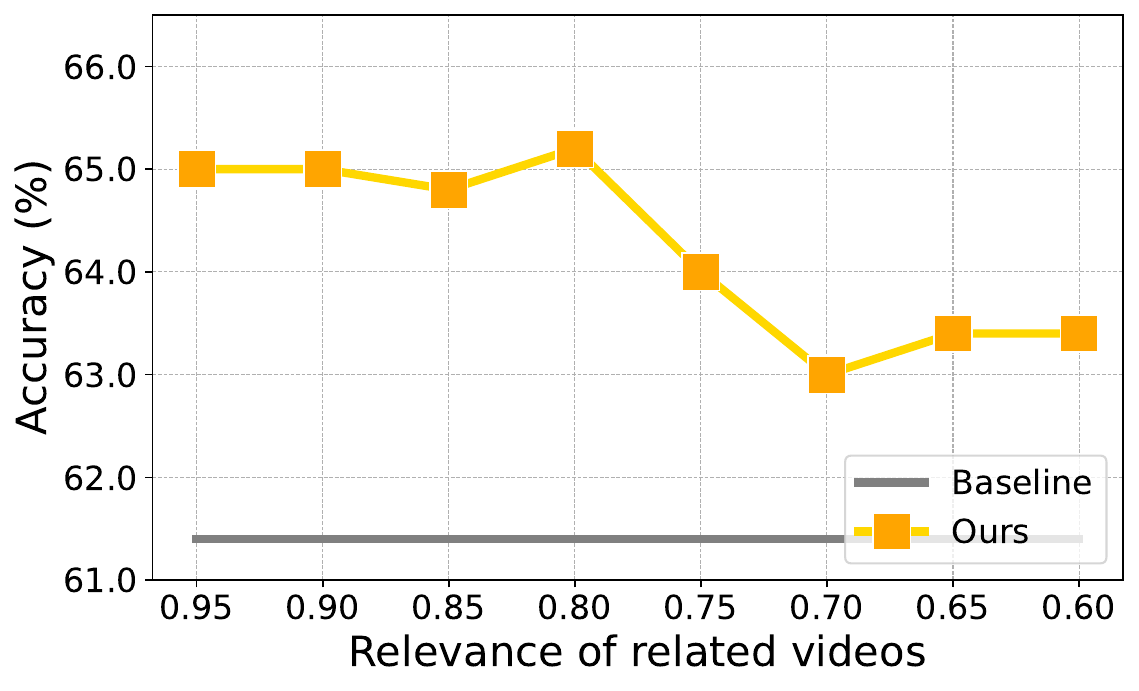} 
        \caption*{(a) LLaVA-OneVision-0.5B.}
    \end{minipage}
    \hspace{10pt}
    \begin{minipage}{0.43\textwidth}
        \centering
        \includegraphics[width=\textwidth]{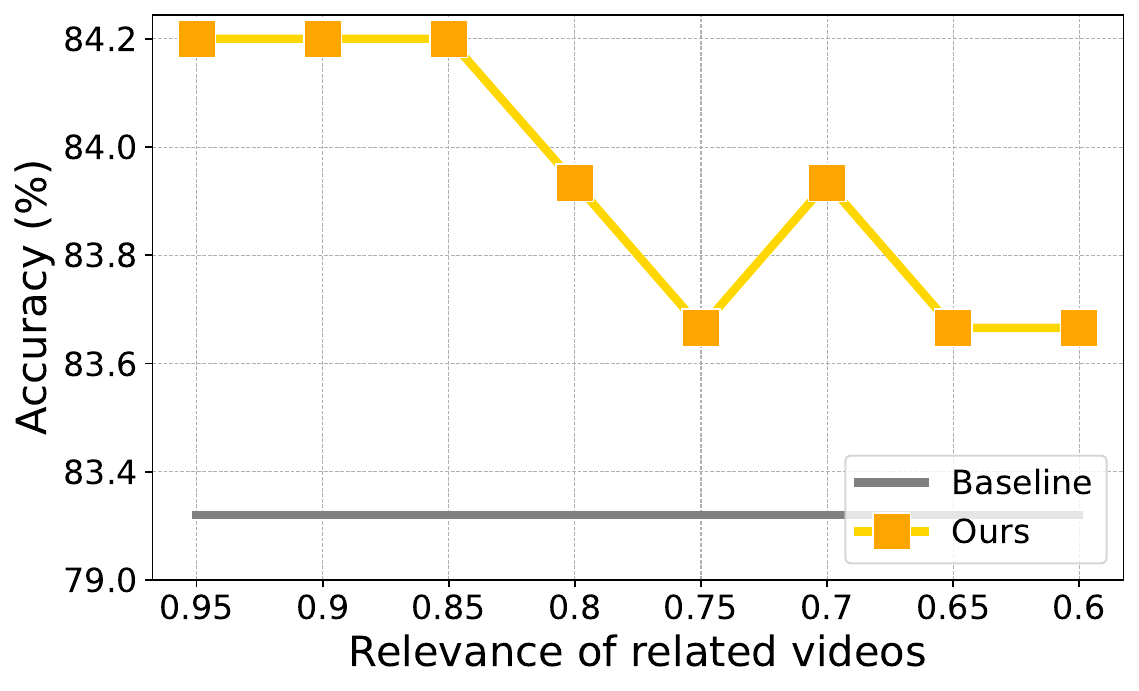} 
        \caption*{(b) LLaVA-Video-7B.}
    \end{minipage}
    \caption{\textbf{Comparative analysis} of accuracy (\%) for NExT-QA across different models under varying relevance of related videos.}
    \label{fig:retrieved_video}
\end{figure*}

\subsection{Discussion on the retrieved video contents}
\label{sec:discussion} 
    \noindent\textbf{How do multiple videos affect the performance?} 
    Multi-video data contributes to achieving more comprehensive reasoning. In this section, we discuss the effect of the number of related videos. 
    As shown in Fig.~\ref{fig:video_num}, increasing the number of retrieved videos from 1 to 8 causes accuracy to initially rise, peaking at 5 videos, before gradually declining. Importantly, this accuracy trend is accompanied by only a marginal increase in the total number of tokens.

    \noindent\textbf{How does video relevance affect the performance?}
    In the above experiments, we arrange the most related videos to the target video within each iteration. In this part, we rearrange the retrieved videos with different lower similarities (measured as cosine similarity between retrieval features) to evaluate how video relevance impacts the performance. As illustrated in Fig.~\ref{fig:retrieved_video}, the reasoning performance decreases as the relevance of the video diminishes, while the performance is still comparable to baseline. 

    \noindent\textbf{How does the video retrieval strategy affect the performance?}
    Intuitively, different retrieval strategies are the potential influencing factor for our collaborative reasoning method. Therefore, for a more detailed discussion, we conduct ablation studies on three video retrieval strategies:
    \begin{itemize}
        \item\textbf{Video vector-based retrieval.}
        This strategy is applicable to most situations. It uses the SigLIP~\cite{zhai2023sigmoid} vision encoder to generate the features of the class tokens from sampled frames, then computes the average feature as each video's feature vector, and constructs the video vector dataset. During inference, $N$ related videos are retrieved based on the highest cosine similarity between feature vectors.
        
        \item\textbf{Caption vector-based retrieval.}
        This strategy applies to video datasets equipped with corresponding captions. It uses the text encoder to extract the feature vector from each video's caption and constructs the caption vector dataset. During inference, $N$ related videos are retrieved by finding the highest cosine similarity between caption feature vectors.

        \item\textbf{Restricted retrieval.}
        This strategy applies to artificially partitioned video datasets. Specifically, during inference, the retrieval process is restricted to videos within the test set, and the retrieval method follows the same procedure as caption vector-based retrieval.
    \end{itemize}

    We implement all three video retrieval strategies and test them on evaluation datasets. The results, presented in Tab.~\ref{tab:retrieval}, indicate that the caption vector-based retrieval achieves the best performance, which can be attributed to the high-quality prompt construction (see Fig.~\ref{fig:caption_prompt}) and the outstanding retrieval capabilities of Qwen3-Embedding~\cite{zhang2025qwen3}. Consequently, we adopt the caption vector-based retrieval strategy in this work. Nonetheless, the other strategies also demonstrate competitive performance, indicating that the reasoning process is only slightly affected by the choice of retrieval strategy. Overall, our framework exhibits robust performance across different video retrieval strategies.

    \noindent\textbf{Conclusion.} 
    Using more relevant videos can lead to better performance, as the key lies in retrieving videos that contain critical information. Incorporating more unrelated or less relevant videos inevitably introduces noise, but our method is capable of filtering out irrelevant information to some extent. Furthermore, our framework exhibits remarkable robustness, maintaining strong performance even when employing different video retrieval strategies.

\begin{figure*}[t]
    \centering
    \includegraphics[width=\textwidth]{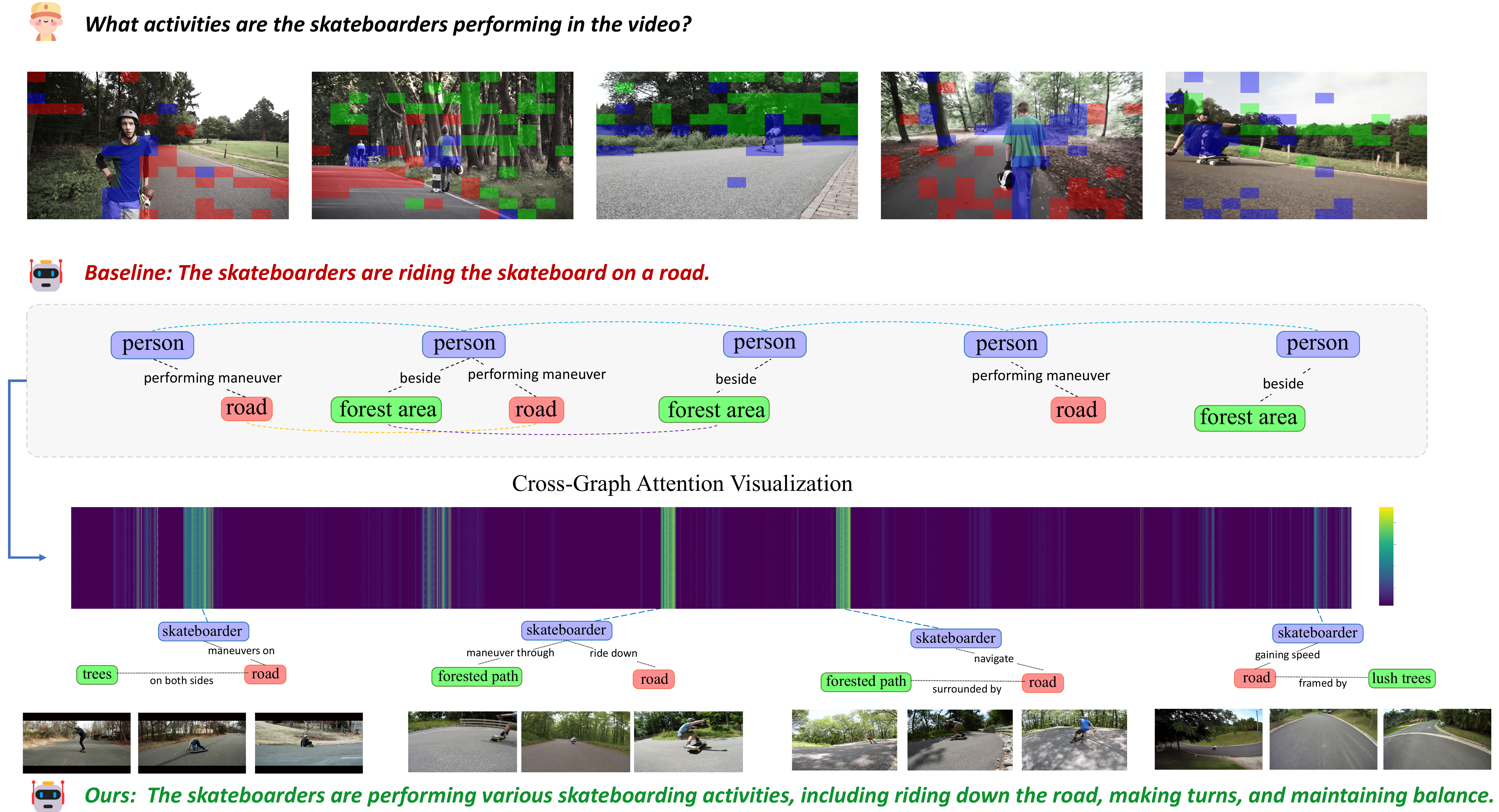}
    \caption{\textbf{Visualization of our structured multi-video collaborative reasoning.} We present a representative video question-answering example from our structured multi-video collaboration pipeline, showcasing the Pooling Attention visualization, structuring results, and the Cross-Graph Attention map, along with the answers generated before and after applying our framework. Within each scene, color patches correspond to triplets with matching colors, highlighting regions of interest identified through Pooling Attention, while the dashed lines indicate the relationships between triplets within the scenes.}
    \label{fig:vis}
\end{figure*}

\subsection{Visualization}
    We visualize the reasoning process of our multi-video collaboration framework in Fig.~\ref{fig:vis}. The query is ``\textit{What activities are the skateboarders performing in the video?}'', which requires high-level and domain-related knowledge. The baseline model failed to provide a detailed response, offering only a generic description. In contrast, our framework represents the video as graph-structured data, preserving key spatio-temporal information.
    In Fig.~\ref{fig:vis}, the color patches correspond to triplets with matching colors, highlighting key regions of interest identified through Pooling Attention. These regions emphasize the most significant visual features necessary for understanding the scene based on their associated triplets. Furthermore, the dashed lines illustrate the relationships between these triplets, demonstrating how structured video representations aggregate and pass based on the relational information across different frames and videos.
    Through the Cross-Graph Attention mechanism shown in Fig.~\ref{fig:vis}, sub-graphs from related videos contribute useful relational structures to the current video’s graph. By fusing sub-graph features from related videos, our model builds a coherent understanding of complex scenarios, leading to an accurate and detailed response.

\begin{figure*}[t]
    \centering
    \includegraphics[width=0.9\linewidth]{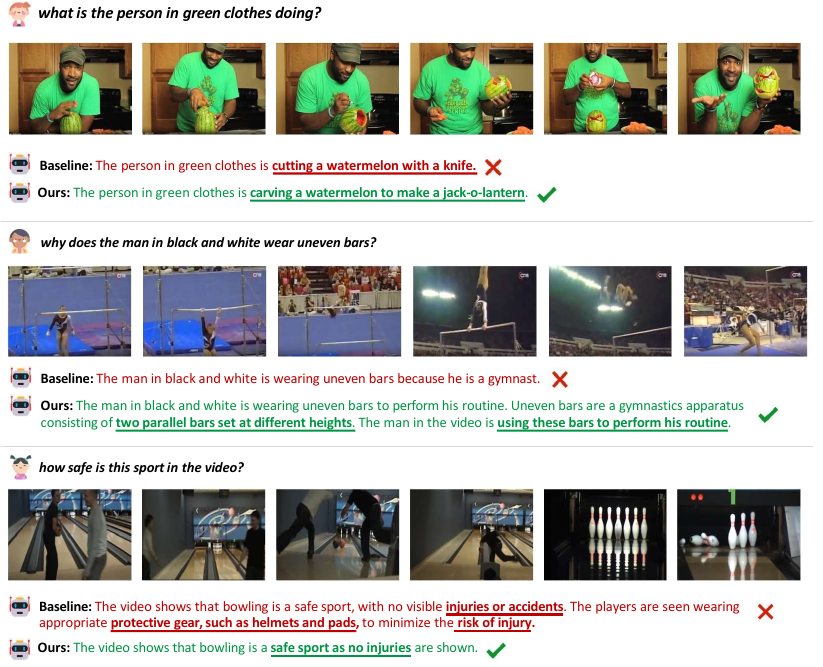}
    \caption{\textbf{Visualization of video question answering examples.}}
    \label{fig:more_vis}
\end{figure*}

\subsection{More Video Question-Answering Results}
    We present additional video question-answering results in Fig.~\ref{fig:more_vis}, which illustrate the strength of our framework in delivering accurate, detailed, and contextually precise answers, surpassing baseline methods.
    
    For instance, our model effectively integrates domain knowledge to correctly interpret unique activities, as demonstrated by accurately identifying that the person in green clothes is ``carving a watermelon to make a jack-o-lantern'' rather than merely ``cutting a watermelon with a knife.'' Similarly, it provides more detailed and context-aware answers about the use of uneven bars by identifying them as ``two parallel bars set at different heights'' and specifying that the man is ``using these bars to perform his routine.'' Furthermore, our framework may alleviate hallucination problems to some extent, as shown in the case of evaluating safety in the bowling video, where it concludes that "bowling is a safe sport as no injuries are shown" instead of generating fabricated details.
    
    These examples demonstrate the robustness and reliability of our structured multi-video collaborative reasoning framework, which outperforms the baseline by providing precise, accurate, and context-aware answers across diverse queries.

\section{Conclusion}
    In this work, we introduce a pioneering framework that enhances video large language models through structured multi-video collaborative reasoning. We first design the Video Structuring Module that models video as a spatio-temporal graph. Then, the Graph Fusion Module integrates related video information into enhanced graph tokens, which are then combined with visual and textual tokens in a multi-video structured prompt for input to the language model.
    Extensive experiments demonstrate the effectiveness and robustness of our approach in comprehending complex video content and answering queries accurately. 
    We hope our work can provide insights for reliable video understanding and stimulate more research interest.

\section{Statements and Declarations}
\subsection{Data Availability Statement}
The original video source and corresponding annotations are legally accessible through the publicly available open-source dataset~\cite{zhang2024videoinstructiontuningsynthetic}. Furthermore, we plan to publicly release both our code for the data processing pipeline and the training dataset utilized for GFM training in the near future.

\subsection{Acknowledgment and Competing Interests}
The paper is supported in part by the National Natural Science Foundation of China (No. 62325109, 62561160155, U21B2013), and in part by the Shanghai `The Belt and Road' Young Scholar Exchange Grant (24510742000).


\bibliography{sn-bibliography}

\end{document}